\documentclass{bmvc2k}

\usepackage[capitalize]{cleveref}
\crefname{section}{Sec.}{Secs.}
\Crefname{section}{Section}{Sections}
\Crefname{table}{Table}{Tables}
\crefname{table}{Tab.}{Tabs.}

\usepackage{wrapfig}
\usepackage{float}
\usepackage{multirow}
\usepackage{colortbl}
\usepackage{tabulary}
\usepackage{etoolbox}
\usepackage{pifont}
\usepackage{makecell}
\usepackage{graphicx}
\usepackage{adjustbox}
\usepackage{amsmath}
\usepackage{amssymb}
\usepackage{booktabs}
\usepackage{rotating}
\usepackage{tablefootnote}
\definecolor{LightCyan}{rgb}{0.8,1,1}
\definecolor{LightGreen}{rgb}{0.6,0.9,0.7}
\definecolor{royalblue(traditional)}{rgb}{0.0, 0.14, 0.4}
\definecolor{royalblue(web)}{rgb}{0.25, 0.41, 0.88}
\usepackage{xcolor, soul}
\sethlcolor{LightGreen}
\colorlet{LightCyan}{LightCyan}
\usepackage{colortbl}
\usepackage{tabularx}
\usepackage{mathtools}
\usepackage{ragged2e}
\def\boxit#1{%
  \smash{\fboxsep=0pt\llap{\rlap{\fbox{\strut\makebox[#1]{}}}~}}\ignorespaces
}

\title{\textsc{GOPro:} \underline{G}enerate and \underline{O}ptimize \underline{Pro}mpts in CLIP using Self-Supervised Learning}

\addauthor{Mainak Singha}{mainaksingha.iitb@gmail.com}{1}
\addauthor{Ankit Jha}{ankitjha16@gmail.com}{1}
\addauthor{Biplab Banerjee}{getbiplab@gmail.com}{1}

\addinstitution{
 Indian Institute of Technology Bombay\\
 Mumbai, India
}
\runninghead{Singha, Jha, Banerjee}{Generate and Optimize Prompts in CLIP}

\begin{document}
\maketitle
\vspace{-0.4cm}
\begin{abstract}
Large-scale foundation models, such as CLIP, have demonstrated remarkable success in visual recognition tasks by embedding images in a semantically rich space. Self-supervised learning (SSL) has also shown promise in improving visual recognition by learning invariant features. However, the combination of CLIP with SSL is found to face challenges due to the multi-task framework that blends CLIP's contrastive loss and SSL's loss, including difficulties with loss weighting and inconsistency among different views of images in CLIP's output space.
To overcome these challenges, we propose a prompt learning-based model called \textsc{GOPro}, which is a unified framework that ensures similarity between various augmented views of input images in a shared image-text embedding space, using a pair of learnable image and text projectors atop CLIP, to promote invariance and generalizability. To automatically learn such prompts, we leverage the visual content and style primitives extracted from pre-trained CLIP and adapt them to the target task.
In addition to CLIP's cross-domain contrastive loss, we introduce a visual contrastive loss and a novel prompt consistency loss, considering the different views of the images. \textsc{GOPro} is trained end-to-end on all three loss objectives, combining the strengths of CLIP and SSL in a principled manner. Empirical evaluations demonstrate that \textsc{GOPro} outperforms the state-of-the-art prompting techniques on three challenging domain generalization tasks across multiple benchmarks by a significant margin. Our code is available at \url{https://github.com/mainaksingha01/GOPro}.
\end{abstract}

%-------------------------------------------------------------------------
% \vspace{-0.2cm}
\section{Introduction}
\label{sec:intro}
Vision-language models (VLMs) or foundational models, such as CLIP \cite{clip} and ALIGN \cite{align}, have recently shown exceptional performance in downstream tasks with zero-shot and few-shot scenarios, by employing image-text pairs contrastively, supported by additional information from hand-crafted text prompts like $\texttt{"a photo of a [cls]"}$. However, prompt engineering can present challenges, and learnable prompting techniques, such as CoOp \cite{coop}, CoCoOp \cite{cocoop}, and CLIP-Adapter \cite{clip-adapter}, have replaced manual prompts, showing better generalization abilities.
For example, CoOp employs learnable parameters to create text prompts, but the generation of prompts under the supervision of visual features \cite{cocoop, prograd, maple} has gained increasing attention. Nevertheless, \textit{learning generalizable prompts by leveraging the pre-trained vision and text backbones of CLIP is still regarded as an open problem, partially due to their overlooking of various image transformations}.

Representation learning is a common approach that involves pre-training a model on a large image dataset, such as ImageNet \cite{krizhevsky2012imagenet}, using a supervised approach, which has shown significant improvements in various downstream tasks. However, self-supervised learning, an unsupervised method, has gained popularity due to its success in language \cite{albert,cert} and recent advancements in vision \cite{ssl_cv1,ssl_cv2}. The main objective of self-supervised learning is to replace laborious supervised pre-training that relies on human annotation by defining an auxiliary task that guides the model towards learning a better embedding space.
Recently, contrastive methods \cite{contrastive1,contrastive2} have emerged as a powerful approach to self-supervised pre-training, outperforming more ad-hoc approaches such as zig-saw solver or rotation prediction \cite{hendrycks2019using}, among others. Typically, contrastive SSL approaches consider a pair of augmentations for the input images and aim to learn identical embeddings for them.

Our aim is to investigate the impact of SSL models in leveraging CLIP for complex class and domain generalization tasks. While this approach is not entirely new, the sole existing model in this regard, SLIP \cite{slip}, has proposed to combine the vision-language contrastive learning of CLIP \cite{clip} with a self-supervision head within a multi-task setup. This approach has shown improved performance and demonstrated that SSL could complement CLIP's objective. However, SLIP's full training of the model from scratch can be resource-intensive, and the use of hand-engineered prompts may not be optimal. Moreover, SLIP does not ensure semantic invariance in the prompt space, which can affect generalization performance.
Therefore, in combining CLIP with SSL, we need to carefully consider two critical factors. \textit{Firstly, we should leverage the pre-trained CLIP backbone while introducing a small set of learnable parameters to learn an SSL-influenced joint image-text embedding space. Secondly, we should replace ad-hoc prompts with learnable prompts to increase generalizability and jointly ensure a better alignment of image-text features.} \\

\noindent \textbf{Our proposed \textsc{GOPro}}: To address the research queries mentioned earlier, we present \textsc{GOPro}, a comprehensive framework that leverages the advantages of contrastive SSL and pre-trained CLIP to generate domain and class generic prompts while enhancing the invariance of embedding space against various image-level geometric and photometric transformations. Our approach ensures that by learning to generate consistent prompts for different augmentations of the original images, better generalization can be achieved.

To accomplish the goals, we propose the introduction of learnable projectors atop CLIP's frozen vision and text encoders. We refer to the text projector as the meta-network, aligning with established literature \cite{cocoop}. To generate augmented views of input images, we leverage popular models such as MoCo v3 \cite{mocov3}, and AugMix \cite{augmix}. On the other hand, in contrast to existing techniques \cite{coop, cocoop} that initialize the prompt learner with hand-crafted tokens like \texttt{a photo of a [CLS]}, we propose to learn prompt distributions per class by exploiting image feature distributions. Our hypothesis is that prompts learned in conjunction with visual features offer superior class generalizability. Additionally, we are interested in differentiating between object content features and style features \cite{li2017demystifying} of images, as demonstrated in \cite{stylip}.
However, unlike \cite{stylip}, which suggests learning individual tokens from style features extracted from each layer of CLIP's vision encoder, we propose concatenating content and style information and utilizing a text projection network to learn prompt token embeddings, as this approach is found to be computationally more efficient.

The projectors are meticulously trained with three primary loss objectives to ensure optimal performance. Firstly, we employ a contrastive loss \cite{simclr1} between MoCo v3 augmentations of the input image, enhancing the invariance of the visual projector. Secondly, we fine-tune the visual and text projectors using a contrastive loss applied to the image-prompt embeddings. Lastly, we introduce a consistency loss that compares the prompt embeddings obtained from the actual input image with those of the augmented views. To improve robustness and optimize uncertainty of the shared embedding space, we consider the augmented views of MoCo v3 along with AugMix, as it is found to boost the classification performance \cite{augmix} given its composition-based more diverse set of image synthesis capabilities. However, while \cite{augmix} focuses on visual feature consistency with AugMix synthesized images, we propose to leverage AuxMix to enforce semantic consistency together with the weak augmentations from MoCo v3 at the prompt space of CLIP, as the final classification is to be carried out there.
We highlight our \textbf{major contributions} as,

\noindent [-] In this paper, we strategically enhance CLIP's prompt learning by using an SSL objective together with the notion of disentangled image-domain-conditioned prompt learning.\\
\noindent [-] Our key contributions involve updating newly-introduced light-weight vision and text projectors atop frozen CLIP using a combination of visual-space SSL contrastive loss, CLIP's image-text contrastive loss, and a novel prompt consistency loss that takes into account the various views of the images. Furthermore, we propose learning the prompt distributions leveraging the multi-scale visual content and style information extracted from CLIP.\\
\noindent [-]  To evaluate the effectiveness of our proposed approach, we conduct extensive experiments across three different settings, including base-to-new class generalization, cross-dataset transfer, and single-source multi-target domain generalization on multiple benchmark datasets (as described in Sec. \ref{sec:dataset}). Our \textsc{GOPro} method significantly outperforms other state-of-the-art comprehensively in all the cases.
  
% \vspace{-0.2cm}
\section{Related Works}
% \vspace{-0.3cm}
\label{sec:literature}
\noindent\textbf{Vision-language models and prompt learning:} 
In general, multimodal learning has been shown to yield better feature learning than unimodal setups. Tasks such as image captioning \cite{shottell}, image retrieval \cite{babenko2014neural}, and visual question answering (VQA) \cite{vqa1} typically require joint visual-semantic supervision. Moving forward, VLMs such as CLIP \cite{clip} have recently gained significant attention. VLMs are trained on large-scale image-text pairs in a contrastive manner to align the visual and textual embeddings. VLMs efficiently transfer the learned vision information via prompt-based zero-shot and few-shot downstream tasks.

Prompt learning is a widely used approach in NLP \cite{petroni2019language} and has recently been applied to visual recognition tasks. The primary aim is to leverage pre-trained language models like BERT \cite{bert} to provide useful information for downstream tasks through semantically meaningful textual prompts. In recent years, research has focused on automating prompt generation to eliminate manual intervention. One such method is AutoPrompt \cite{shin2020autoprompt}, which examines tokens with the most significant gradient changes in the label likelihood. Meanwhile, CoOp \cite{coop} optimizes prompts by fine-tuning CLIP for few-shot image classification. CoCoOp \cite{cocoop} suggests learning conditional prompts based on image features, which can improve CoOp's generalization capability. Crisply, CoCoOp and ProGrad \cite{prograd} generate prompts from high-level visual features and optimize the generated context tokens. Besides, prompt distribution learning (PDL) \cite{pdl} proposes optimizing multiple sets of prompts and APPLeNet \cite{applenet} has demonstrated significant domain generalization performance using multi-scale features within remote sensing images. In the other hand, AD-CLIP \cite{adclip} has exhibited notable results in domain adaptation by harnessing visual tokens within the prompt space. While these methods focus on image data, Video Prompt Learning (VPL) \cite{vpl} proposed leveraging foundation models for video data. Finally, Self-supervised Learning with Inter-modality Prompts (SLIP) \cite{slip} proposed supplementing the contrastive learning of CLIP with an SSL objective in a multi-task setup. \textit{However, the SSL objective is applied only to the visual branch and is disjoint from the semantic branch, thus not leveraging the multi-modal aspect of CLIP comprehensively. In contrast, we ensure that the SSL objective improves the learning of both the visual and semantic projectors, resulting in enhanced generalization.}

\textit{Our approach to prompt learning differs from the literature \cite{coop, cocoop, stylip} in the way we utilize visual information. While we draw inspiration from StyLIP \cite{stylip} in the idea of disentangling image content and style information, we diverge significantly from \cite{stylip} in how we leverage the visual information to initialize prompt tokens. In StyLIP, each prompt token is learned solely from style information extracted from a specific layer of CLIP's vision encoder, which limits its ability to handle prompts of varied context lengths. In contrast, our approach combines multi-scale content information with global style information from the final vision encoder layer and subsequently learns prompt tokens through a shared meta-network. This offers more flexibility in the length of prompts, allowing for prompts of different context lengths to be effectively learned and utilized in our approach.}

\vspace{0.2cm}
\noindent\textbf{Self-Supervised Learning:}
Self-supervised learning (SSL) is a technique that aims to learn high-quality visual representations from unlabeled images without additional human supervision. Advancements in SSL have made it possible to narrow the gap between supervised and unsupervised representations, as evaluated in downstream tasks \cite{contrastive1,unsup1,unsup2,moco}. One popular approach is contrastive learning \cite{gutmann2010noise,wang2020generalizing}, which aims to embed augmented views of a given image closely in feature space while pushing away other images in the same batch \cite{contrastive1} or using a memory bank \cite{moco}. Other methods focus on retrieving more informative positive examples during training that exhibit more natural image variation than simple artificial augmentations \cite{ayush2021geography,oord2018representation}. Some contrastive variants even report strong performance without negative examples \cite{chen2021exploring,grill2020bootstrap}. \cite{cole2022does} showed that self-supervised training on ImageNet \cite{krizhevsky2012imagenet} is still highly effective even when using less than $25\%$ of the unlabeled images during training, outperforming supervised pre-training. \textit{As opposed to these approaches, we are keen on improving the invariance of VLMs by supplementing an SSL task, which is found to be beneficial for domain and class generalization tasks.}

% \vspace{-0.2cm}
\section{Problem Definition and Proposed Methodology}

Let $\mathcal{D}_s = \{\mathcal{D}_s^i\}_{i=1}^{n}$ denote $n$ source domains, each with input data $x^i \in\mathcal{X}^i$ and corresponding label space $y^i\in\mathcal{Y}_{Seen}$. It's important to note that the probability distribution of each domain, $P(\mathcal{D}_s^i)$, may differ for all $i\in{1,\cdots,n}$. During training, we use the labels $\mathcal{Y}_{Seen}$ from $\mathcal{D}_s$, while during testing, we use $\mathcal{Y}_{Unseen}$ from a target test domain $\mathcal{D}_t$ with $\mathcal{P}(\mathcal{D}_t) \neq \mathcal{P}(\mathcal{D}_s^i)$, $\forall i\in{1,\cdots,n}$. For base-to-new class generalization, we set $\mathcal{Y}_{Seen} \cap \mathcal{Y}_{Unseen} = \emptyset$. In contrast, for domain generalization (DG), we consider single-source DG and assume that the label sets for both domains are identical ($\mathcal{Y}_{Seen} \cap \mathcal{Y}_{Unseen} = \mathcal{Y}_{Seen} \cup \mathcal{Y}_{Unseen}$). Finally, for the across-dataset DG, there may or may not be some overlap between $\mathcal{Y}_{Seen}$ and $\mathcal{Y}_{Unseen}$.

% \vspace{-0.2cm}
\subsection{Explaining the working principles of \textsc{GOPro}}

This section provides an elaborate description of the architecture and training methodology of \textsc{GOPro}. Specifically, \textsc{GOPro} utilizes the visual backbone of CLIP, denoted as $f_v$, and the text encoder, denoted as $f_t$. Let's assume $f_v$ consists of $L$ encoder layers, and the feature responses for layer $l \in [1,L]$ are denoted as $f_v^l$. Additionally, we introduce the following learnable units in \textsc{GOPro}: a vision projector $P_v$ in the visual space, and text projectors $\rho_f$, $\rho_{Au}$, and $\rho_{mo}$ for obtaining token embeddings from visual features from the original and the augmented images (MoCo v3 and AuxMix generated), respectively. For simplicity, we consider $\rho = \rho_f = \rho_{Au} = \rho_{mo}$ in our experiments. The goal is to train $(\rho, P_v)$ using image-prompt pairs, such that the learned shared embedding space is generalizable, discriminative, and invariant to image transformations. To achieve this, we propose the following novel considerations:
i) A novel prompt learning scheme that leverages multi-scale visual content features extracted from $f_v$, along with image style information in terms of the instance-wise feature statistics from the $L^{th}$ layer of $f_v$ is introduced.
ii)   For training $P_v$ using dataset $\mathcal{D}_s$, we deploy the MoCo v3 augmentation technique and use the contrastive formulation of SimCLR \cite{simclr1}, following SLIP \cite{slip}.
iii)  To train $P_v$, we consider the MoCo v3 augmentation $x_1$ for the original image $x$ and the contrastive loss is realized for $(x,x_1)$. On the other hand, we consider an AugMix-generated image $x_2$ for $x$.
We then carry out image-text contrastive loss for $(x,y)$, while simultaneously ensuring that $f_t(x,y) \sim f_t(x_1,y) \sim f_t(x_2,y)$.

In this way, we fully leverage the rich representation space of CLIP and smoothly adapt the model for the downstream DG tasks with few training samples. 

\begin{figure}
    \centering
    \includegraphics[scale=0.6]{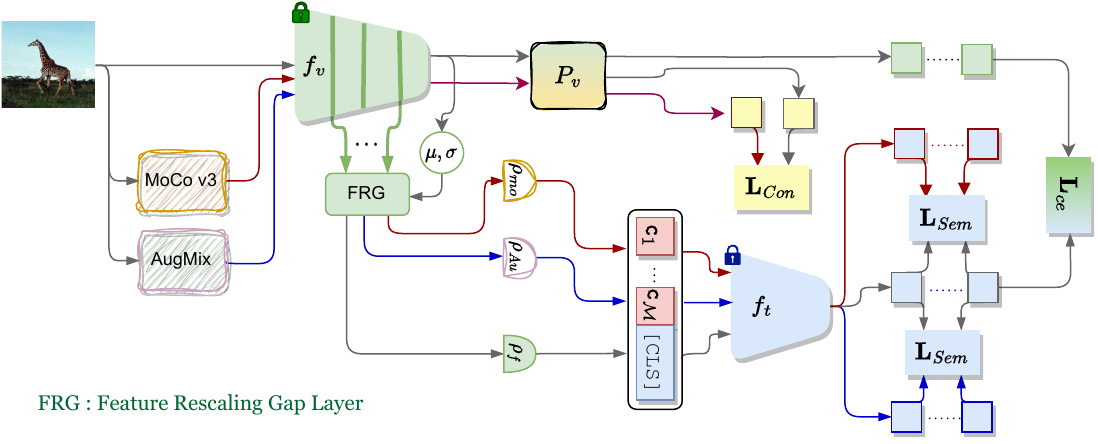}
    \vspace{0.2cm}
    \caption{\textcolor{royalblue(traditional)}{\textbf{The design of \textsc{GOPro}} entails utilizing the fixed image $f_v$  and text encoders $f_t$ of CLIP. In addition, \textsc{GOPro} incorporates several distinct trainable meta-networks that generate tokens for the original image and augmented images created by MoCo v3 \cite{mocov3} and AugMix \cite{augmix}, denoted as $\rho_{f}$, $\rho_{mo}$ and $\rho_{Au}$ respectively. To rescale the features from intermediate layers of $f_v$, the architecture employs a combination of feature rescaling and the global average pooling (GAP) operation, which we collectively referred to as the FRG layer.}}
    \label{fig:GOPro}
    \vspace{-0.3cm}
\end{figure}

\vspace{0.2cm}
\noindent \textbf{Image content and style driven prompt generation:} In our approach, we aim to generate prompts from the visual features $f_v(x)$ by disentangling the content and style components. To achieve this, we utilize multi-scale content features obtained from different layers of $f_v$, denoted as $\hat{F}(x) = [\hat{f}_v^1(x); \cdots ;\hat{f}_v^L(x)]$ after concatenation. The key idea behind this multi-scale representation is that $\hat{F}(x)$ captures low, mid, and high-level features in its different layers, making it more transferable compared to considering only high-level semantic features \cite{stylip}.
Similarly, we represent the style features using instance-wise feature statistics, namely channel-wise mean and standard deviation, calculated from the $L^{th}$ layer of $f_v$. Precisely, $\bar{F}(x) = [\vec{\mu_{L}}(x); \vec{\sigma_{L}}(x)]$ denotes the style vector. The prompt token initialization for image $x$ is then represented as $F(x) = [\hat{F}(x); \bar{F}(x)]$, rescaled through feature rescaling gap (FRG) layer and it is further adapted to the distribution of $\mathcal{D}$ using $\rho$. In \textsc{GOPro}, $\rho$ takes the structure of a single encoder and $\mathcal{M}$ decoders, where $\mathcal{M}$ defines the context length. This way, we learn $\mathcal{M}$ distinct tokens given $F(x)$. The prompt for class $y$ can be derived as: $\text{Pr}_y(x,y) = [c_1; c_2; \cdots; c_{\mathcal{M}};[CLS_y]]$, where $c_m$ is the output from the $m^{th}$ decoder of $\rho$, and $CLS_y$ is the semantic embedding for the class $y$. In the \texttt{Supplementary}, we discuss how our approach differs from the existing prompt learning techniques \cite{stylip, coop, cocoop}.

\vspace{0.2cm}
\noindent \textbf{Visual self-supervised objective:} We explore the synergistic combination of the SimCLR SSL objective with augmentations obtained from MoCo v3, as reported in the literature \cite{slip}, to train $P_v$. The standard normalized temperature-scaled cross-entropy loss formulation, denoted as $\mathbf{L}_{Con}$ \cite{simclr1}, is employed to maximize the \texttt{cosine} similarity ($\delta$) between an original image $x$ and its augmented view $x_1$.

\vspace{0.2cm}
\noindent \textbf{Image-text mapping and prompt consistency objective:} Our approach utilizes contrastive learning to map visual and text feature embeddings into a shared embedding space. To compute the class posterior probability of an input $x$ belonging to class $y$, we employ the following definition, where $\tau$ represents the temperature hyper-parameter. We consider the output of $P_v$ as the visual embeddings since this space promotes visual invariance.
\begin{equation}
p(y|x) = \frac{ \exp(\delta(P_v(f_v(x)), f_t(\text{Pr}_{y}(F(x)))/\tau))}{\sum_{k=1}^{|\mathcal{Y}_{Seen}|}\ \exp(\delta(P_v(f_v(x)), f_t(\text{Pr}_{y_k}(F(x)))/ \tau))}
\end{equation}

Subsequently, the cross-entropy loss ($\mathbf{L}_{ce}$) is computed between the prediction probabilities of each input image and their corresponding class labels as follows:
\begin{equation}
    \mathbf{L}_{ce} = \underset{P_v, \rho}{\arg\min} \underset{(x,y) \in \mathcal{P}(\mathcal{D}_s)}{\mathbb{E}}  - \sum_{k=1}^{\mathcal{Y}_{Seen}} y_{k} log(p(y_k|x)_{f_v,f_t})
\label{lce}
\end{equation}

We emphasize the importance of consistent prompt embeddings obtained from various augmentations applied to the input image. This is crucial in achieving semantic invariance, complementing the visual invariance ensured by $\mathbf{L}_{Con}$. To achieve this, we employ two augmentations per image $x$: one generated by MoCo v3, which applies geometrical transformations to $x$, and the other generated by AugMix. AugMix is particularly useful in scenarios where the data distribution encountered during deployment may differ from the training distribution, such as when images are captured with different cameras. AuxMix has been demonstrated to significantly improve generalization performance without necessitating changes to the underlying model \cite{augmix}.

Our approach to achieving semantic consistency involves incorporating distillation losses based on an $\ell_2$-norm distance measure. Here, the prompt embedding of the original image $x$ serves as the teacher, while the prompt embeddings of the two augmentations, Moco v3 ($x_1$) and AugMix ($x_2$), serve as the students. The loss is defined as,

\vspace{-0.5cm}
\begin{equation}
    \centering
    \label{eq:3}
    \begin{aligned}
    \mathbf{L}_{Sem} = \underset{P_v, \rho}{\text{argmin}} \underset{\mathcal{P}(\mathcal{D}_s)}{\mathbb{E}} ||f_t(\text{Pr}_y(\rho(F(x)))) & - f_t(\text{Pr}_y(\rho(F(x_1))))||_2 \\
    & + ||f_t(\text{Pr}_y(\rho(F(x))))- f_t(\rho(\text{Pr}_y(F(x_2))))||_2
    \end{aligned}
\end{equation}

\noindent \textbf{Total loss for training and inference:} We train the model with respect to all the losses mentioned above, where,
\vspace{-0.25cm}
\begin{equation}
\centering
\mathbf{L}_{Total} = \mathbf{L}_{Sem} + \mathbf{L}_{ce} + \mathbf{L}_{Con}
\end{equation}

During inference, we generate all the class prompts given $\mathcal{Y}_{Unseen}$ for a given $x_t$, and the $y \in \mathcal{Y}_{Unseen}$ maximizing $p(y|x_t)$ is selected.

\section{Experimental Evaluations}
% \vspace{-0.1cm}
\noindent \textbf{Dataset descriptions:}
\label{sec:dataset}
We evaluate \textsc{GOPro} on 11 image recognition datasets for base-to-new class generalization and cross-dataset transfer, following the procedure described in CoOp \cite{coop}. The datasets include ImageNet \cite{krizhevsky2012imagenet}, Caltech101 \cite{caltech} for generic object classification, OxfordPets \cite{oxford}, StanfordCars \cite{stanford}, Flowers102 \cite{flower}, Food101 \cite{food}, and FGVCAircraft \cite{aircraft} for fine-grained classification, SUN397 \cite{sundataset} for scene recognition, UCF101 \cite{ucf} for action recognition, DTD \cite{dtd} for texture classification, and EuroSAT \cite{helber2019eurosat} for satellite imagery recognition. For domain generalization experiments, we employ ImageNet as the source dataset and four other ImageNet variants as target datasets, namely ImageNetV2 \cite{imgv2}, ImageNet-Sketch \cite{img_sketch} - It consists of 50000 images, 50 images for each of the 1000 ImageNet classes, ImageNet-A \cite{img_A}, and ImageNet-R \cite{img_R}. 

% \vspace{0.1cm}
\noindent \textbf{Architecture Details:}
$\rho$ is implemented as a two-layer bottleneck network followed by Linear-ReLU-Linear, where the hidden layer is expanded to the number of context tokens. On the other hand, $P_v$ follows a single-layer MLP structure with a batch normalization layer. $f_v$ and $f_t$ are realized using CLIP's pre-trained transformer backbones.\\
% \vspace{0.1cm}
\noindent \textbf{Training and evaluation protocols:}
To train \textsc{GOPro}, we utilize the stochastic gradient descent (SGD) optimizer \cite{robbins1951stochastic} for 50 epochs and apply scheduling to avoid local minima. During training, we employ 16 shots (samples per class) with a batch size of 4 and ViT-B/16 as the image encoder backbone. The text prompts are initialized using \texttt{"a photo of a [CLS]"}, indicating a context length, $\mathcal{M}$=4, as per previous literature \cite{cocoop}. We report the average \texttt{top-1} accuracy from three runs of the model. 

% \vspace{-0.2cm}
\subsection{Comparisons to the state-of-the-art}
% \vspace{-0.1cm}
\noindent \textbf{Baselines \& competitors}: In our performance evaluation of \textsc{GOPro}, we compare it to existing methods from the prompting literature using CLIP. As our baselines, we utilize Zero-shot CLIP \cite{clip} and the SSL-based SLIP \cite{slip} models, respectively. Additionally, we compare our model with prompt learning techniques, such as CoOp \cite{coop}, CoCoOp \cite{cocoop}, MaPLe \cite{maple}, and \textsc{StyLIP} \cite{stylip}, using ViT-B/16 backbone.

\begin{wraptable}{r}{0.5\textwidth}
\vspace{-0.75cm}
\begin{center}
\caption{\textcolor{royalblue(traditional)}{Comparison of \textsc{GOPro} with state-of-the-art methods on B2N generalization on the average metrics over 11 visual recognition datasets. HM represents the harmonic mean.}}\label{tab:B2N}
\scalebox{0.8}{
\begin{tabular}{lcccc}\\
\toprule
\rowcolor{gray!20} {\textbf{Method}} & {Base} & {Novel} & {HM} \\ 
        \midrule
          \cellcolor[gray]{0.9} CLIP \cite{clip} &69.34 &74.22 &71.70  \\

      \cellcolor[gray]{0.9} SLIP \cite{slip}&69.77 &74.28 &71.96 \\

      \cellcolor[gray]{0.9}CoOp \cite{coop}&82.69 &63.22 &71.66 \\

      \cellcolor[gray]{0.9}CoCoOp \cite{cocoop}&80.47 &71.69 &75.83\\

      \cellcolor[gray]{0.9}MaPLe \cite{maple}&82.28 &75.14 &78.55 \\

      \cellcolor[gray]{0.9}\textsc{StyLIP} \cite{stylip}& 83.22 &75.94 &79.41 \\
      
      \cellcolor{green!15}\textsc{GOPro}&{\textbf{84.21}} &
{\textbf{77.32}} &{\textbf{80.62}}\\
\bottomrule
\end{tabular}}
\end{center}
\vspace{-0.6cm}
\end{wraptable} 

\vspace{0.2cm}
\noindent\textbf{Base-to-New (B2N) class generalization:} Table \ref{tab:B2N} showcases the experimental results for B2N class generalization averaged over 11 fine-grained and coarse-grained datasets. The harmonic mean (H) between the classification accuracies of the Base and Novel classes is computed. To ensure fairness, we randomly and equally divide the datasets into two groups, defining the base and novel classes for training and testing, respectively. Given that \textsc{GOPro} is a self-supervised prompt learning method, we pay particular attention to SLIP's self-supervised zero-shot approach. Remarkably, \textsc{GOPro} achieves superior generalization scores, surpassing SLIP by a significant margin of 14.44\% on seen classes and 3.04\% on unseen classes across all datasets on average. Furthermore, we compare \textsc{GOPro} with recent context optimization-based methods, demonstrating its superiority over MaPLe and \textsc{StyLIP} by 2.07\% and 1.21\% on average, respectively. 

% The detailed results are mentioned in the \texttt{Supplementary}.

\vspace{0.2cm}
\noindent\textbf{Cross-Dataset (CD) generalization:} Table \ref{tab:CDT} showcases the evaluation results of \textsc{GOPro} on the CD setup, where the model is trained on the ImageNet dataset (source domain) and zero-shot inference is performed on the remaining ten datasets (target domains). Remarkably, \textsc{GOPro} surpasses the target classification performance of CLIP and SLIP by substantial margins of 2.63\% and 2.55\%, respectively. Moreover, \textsc{GOPro} outperforms MaPLe and \textsc{StyLIP} by 1.56\% and 0.48\%, respectively, on average. These results demonstrate that \textsc{GOPro} effectively mitigates the generalization gap for diverse domains and classes.\\

% \vspace{-0.15cm}
\begin{table*}[!ht]
% \vspace{-0.3cm}
% \scriptsize{
    \centering
    \caption{\textcolor{royalblue(traditional)}{Comparison of \textsc{GOPro} with the prompt benchmark methods for generalization across datasets. We train the model on ImageNet using $16$-shots with CLIP ViT-B/16 and test on $10$ other datasets.}}
    \vspace{0.2cm}
    \scalebox{0.65}{
    \begin{tabular}{lccccccccccccc}
    \toprule
       \rowcolor{gray!20} & \multicolumn{1}{c}{\textbf{Source}} & \multicolumn{11}{c}{\textbf{Target}}  \\ \cmidrule(lr){2-2} \cmidrule(lr){3-13}
        
       \rowcolor{gray!20} \multirow{-2}{*}{\textbf{Method}} & {ImgNet.} & {C101} & {Pets} & {Cars} & {Flowers} & {Food} & {Aircraft} & {Sun397} & {DTD} & {EuroSAT} & {UCF101} & {Average} \\ 
        \midrule
        \cellcolor[gray]{0.9}CLIP \cite{clip} & 66.73 & 93.31 & 89.10 & 65.64 & 70.73 & 85.86 & 24.72 & 62.58 & 44.39 & 48.28 & 67.72 & 65.23 \\ 

        \cellcolor[gray]{0.9}SLIP \cite{slip}& 68.01& 93.52& 89.23& 65.42& 70.55& 85.92& 25.04& 62.74& 44.16& 48.61& 67.93& 65.31 \\ 
        
        \cellcolor[gray]{0.9}CoOp \cite{coop} & 71.51 & 93.70 & 89.14 & 64.51 & 68.71 & 85.30 & 18.47 & 64.15 & 41.92 & 46.39 & 66.55 & 63.88 \\ 
        \cellcolor[gray]{0.9}CoCoOp \cite{cocoop}& 71.02 & 94.43 & 90.14 & 65.32 & 71.88 & 86.06 & 22.94 & 67.36 & 45.73 & 45.37 & 68.21 & 65.74\\
        \cellcolor[gray]{0.9}MaPLe \cite{maple}& 70.72&	93.53&	90.49&	65.57&	72.23&	86.20&	24.74&	67.01&	46.49&	48.06&	68.69&	66.30\\

        \cellcolor[gray]{0.9}\textsc{StyLIP} \cite{stylip}& 72.30& \textbf{95.45}& 91.60& 67.09&	72.36& \textbf{88.60} & 25.21& 68.11&	47.86&	48.22&	69.30	&67.38\\
        
      % GOPro$^{*}$ & 71.80	&94.69&	90.83&	66.48&	71.30&	86.19&	23.85&	66.90&	47.71&	47.80&	68.51&	66.43\\
\cellcolor{green!15}\textsc{GOPro}&\textbf{73.27}&94.81&	\textbf{92.73}&	\textbf{68.67} & \textbf{72.60}&	87.74&	\textbf{25.85}&	\textbf{68.70}&	\textbf{48.04}&	\textbf{49.43}&	\textbf{69.98}	&\textbf{67.86}\\
\bottomrule
        % GOPro&  &  & & & & & & & & & & \\\hline
        
    \end{tabular}}
    \label{tab:CDT}
    % \vspace{-0.2cm}
\end{table*}
%%%%%%%%%%%%%%%%%%%%%%%%%%%%%%
%%%%%%%%%%%%%%%%%%%%%%%%%%%%%%%%%%%%%%%%%%%%%%%%%%$
\begin{figure}
    \centering
    \includegraphics[scale=0.45]{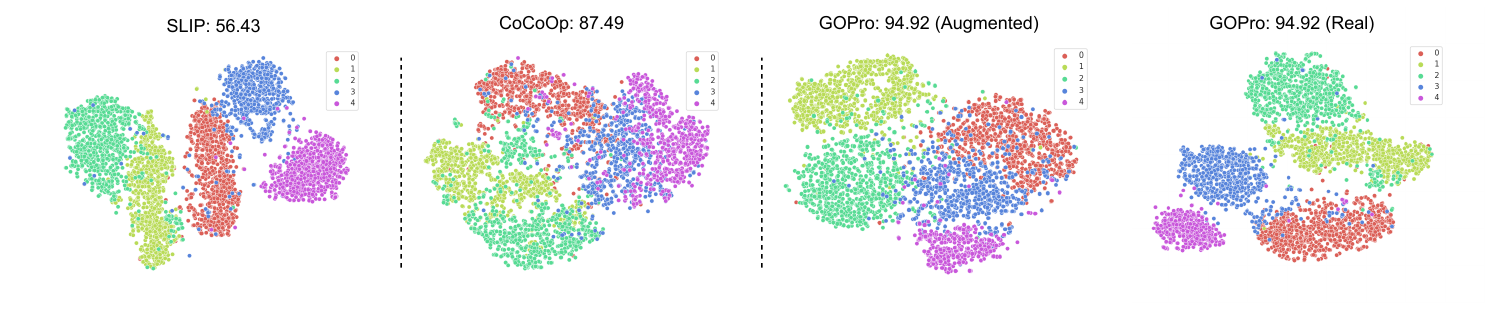}
    % \vspace{0.045cm}
    \caption{\textcolor{royalblue(traditional)}{The t-SNE visualizations of visual embeddings from SLIP, CoCoOp and our proposed \textsc{GOPro}, on the base classes of Eurosat dataset. \textsc{GOPro} archives better discriminativeness.}}
    \label{fig:tsne}
    %\vspace{-0.6cm}
\end{figure}
\vspace{0.2cm}
\noindent\textbf{Domain generalization (DG):}
We have conducted experiments to evaluate the generalization performance of \textsc{GOPro} on a single-source multi-target (SSMT) DG setup. Unlike the CD setting discussed earlier, we only consider the common classes across all datasets, as SSMT is a closed-set setting. The model is trained on the ImageNet dataset and evaluated on its domain variant datasets. Comparison results with state-of-the-art (SOTA) methods and \textsc{GOPro} are presented in Table \ref{tab:DG}.

The results demonstrate that \textsc{GOPro} has outperformed all competitors in the source domain, with a minimum margin of 0.97\%. Additionally, \textsc{GOPro} outperforms other methods in the target domains as well, with minimum margins of 1.07\%, 1.09\%, and 1.49\% in ImageNetV2, ImageNet-A, and ImageNet-R, respectively, except for ImageNet-Sketch, where \textsc{StyLIP} achieves the best performance.

% \vspace{-0.15cm}
\begin{table*}[!ht]
% \vspace{-0.3cm}
% \scriptsize{
    \centering
    \caption{\textcolor{royalblue(traditional)}{Comparison of \textsc{GOPro} with the prompt benchmark methods for domain generalization across datasets. We train the model on ImageNet using $16$-shots with CLIP ViT-B/16 and test on $4$ other datasets.}}
    \vspace{0.2cm}
    \scalebox{0.65}{
    \begin{tabular}{lcccccc}
    \toprule
       \rowcolor{gray!20} & \multicolumn{1}{c}{\textbf{Source}} & \multicolumn{4}{c}{\textbf{Target}}  \\ \cmidrule(lr){2-2} \cmidrule(lr){3-6}
        
       \rowcolor{gray!20} \multirow{-2}{*}{\textbf{Method}} & {ImageNet} & {ImageNetV2} & {ImageNet-Sketch} & {ImageNet-A} & {ImageNet-R} \\ 
        \midrule
         \cellcolor[gray]{0.9}CLIP \cite{clip} & 66.73 & 60.83 & 46.15 & 47.77 & 73.96  \\ 
         \cellcolor[gray]{0.9}SLIP \cite{slip}& 68.01& 61.12& 46.35& 47.54& 73.88 \\
        \cellcolor[gray]{0.9}CoOp \cite{coop} & 71.51 & 64.20 & 47.99 & 49.71 & 75.21  \\ 
        \cellcolor[gray]{0.9}CoCoOp \cite{cocoop}& 71.02 & 64.07 & 48.75 & 50.63 & 76.18 \\
        \cellcolor[gray]{0.9}MaPLe \cite{maple}& 70.72&	64.07&	49.15&	50.90&	76.98\\
        \cellcolor[gray]{0.9}\textsc{StyLIP} \cite{stylip}& 72.30&	64.28&	\textbf{50.83}&	51.14&	76.53\\
        
      % GOPro$^{*}$ & 71.80	&94.69&	90.83&	66.48&	71.30&	86.19&	23.85&	66.90&	47.71&	47.80&	68.51&	66.43\\
\cellcolor{green!15}\textsc{GOPro}&\textbf{73.27}&\textbf{65.35}& 50.36&	\textbf{52.23}&	\textbf{78.02}\\
\bottomrule
        % GOPro&  &  & & & & & & & & & & \\\hline
        
    \end{tabular}}
    \label{tab:DG}
    \vspace{-0.6cm}
\end{table*}

%%%%%%%%%%%%%%%%%%%%%%%%%%%%%% 

% \vspace{0.1cm}
\subsection{Ablation Analysis}
\noindent \textbf{t-SNE visualization}: We present a t-SNE \cite{tsne} visualization of the image embeddings in Figure \ref{fig:tsne}, generated by the visual features of the original and augmented images. We compare them with SLIP \cite{slip} and CoCoOp \cite{cocoop} on the EuroSAT dataset for the B2N generalization task. The visualization clearly demonstrates that \textsc{GOPro} has better clustering of each class, while the cluster points of many classes get overlapped in CoCoOp.

\begin{wrapfigure}{r}{0.5\textwidth}
% \vspace{-0.3cm}
  \centering
    \includegraphics[scale=0.3]{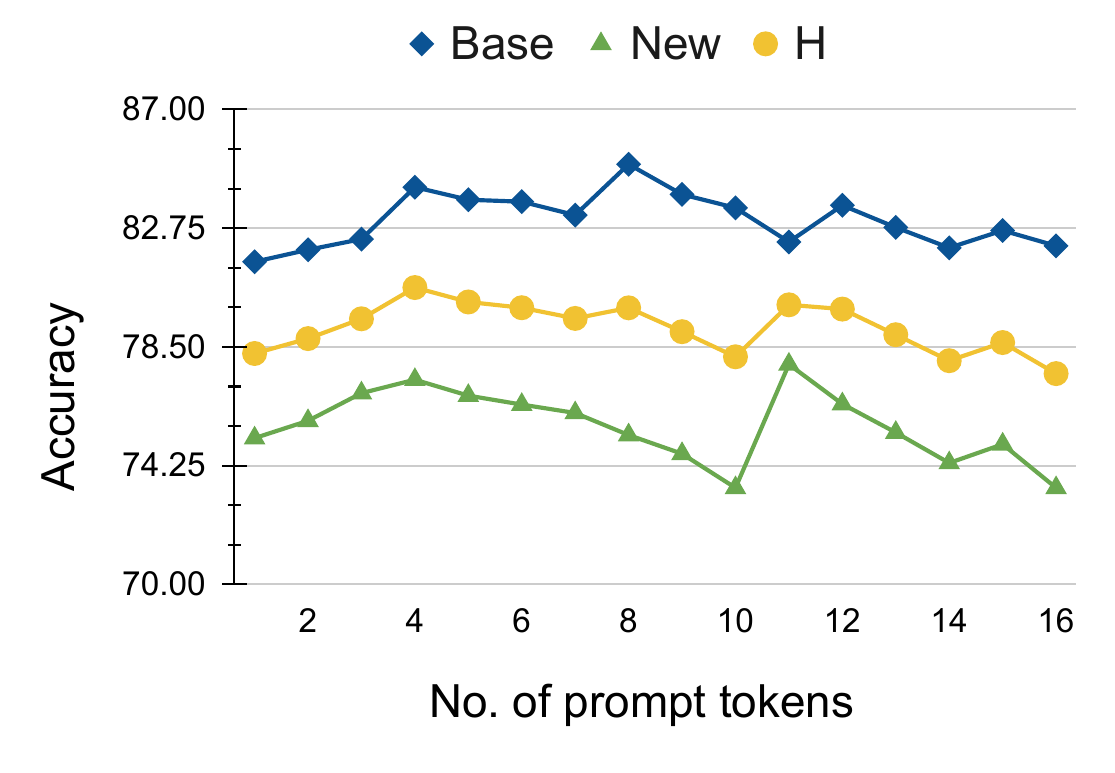}
    % \vspace{0.2cm}
    \caption{\textcolor{royalblue(traditional)}{Comparison of results of \textsc{GOPro} with different numbers of prompt tokens in B2N generalization setup.}}
    \label{fig:tokens_ablation}
\vspace{-0.1cm}
\end{wrapfigure}
% \vspace{-0.2cm}
\vspace{0.1cm}
\noindent \textbf{Sensitivity to context length for B2N generalization}: We have conducted tests on \textsc{GOPro} using varying prompt tokens ($\mathcal{M}$) ranging from 1 to 16. Instead of manual prompt initialization, we randomly initialize prompts from the context. In Figure \ref{fig:tokens_ablation}, we present the average performance of \textsc{GOPro} on 11 datasets in the B2N generalization task with different context lengths. The results indicate that \textsc{GOPro} performs well on base classes with eight tokens and new classes with 11 tokens. However, for better overall generalization, it performs best with four tokens, considering the harmonic mean of both. Interestingly, \textsc{GOPro} achieves almost the same accuracy as random initialization with a context length of 4 in manual initialization of \texttt{"a photo of a"}, as shown in Table \ref{tab:B2N}.

\begin{wraptable}{l}{0.5\textwidth}
\vspace{-0.5cm}
\begin{center}
\caption{\textcolor{royalblue(traditional)}{Ablation study of \textsc{GOPro} with different losses in B2N generalization setup.}\label{tab:loss}}
\scalebox{0.8}{
\begin{tabular}{lcccc}\\
\toprule
\rowcolor{gray!20} {Loss} & {Base} & {Novel} & {HM} \\ 
        \midrule
      \cellcolor[gray]{0.9}$\mathbf{L}_{ce}$ &81.34 &72.16 &76.48  \\

      \cellcolor[gray]{0.9}$\mathbf{L}_{ce}$ + $\mathbf{L}_{Con}$ &82.15 &75.02 &78.42 \\

      \cellcolor[gray]{0.9}$\mathbf{L}_{ce}$ + $\mathbf{L}_{Sem}(x_1)$ &83.23 &74.64 &78.70 \\

      \cellcolor[gray]{0.9}$\mathbf{L}_{ce}$ + $\mathbf{L}_{Sem}(x_2)$ &81.65 &73.97 &77.62 \\

      \cellcolor[gray]{0.9}$\mathbf{L}_{ce}$ + $\mathbf{L}_{Sem}(x_1 + x_2)$ &83.87 &75.15 &79.27 \\
      
      \cellcolor{green!15}$\mathbf{L}_{ce}$ + $\mathbf{L}_{Sem}$ + $\mathbf{L}_{Con}$ &{\textbf{84.21}} & {\textbf{77.32}} &{\textbf{80.62}}\\
\bottomrule
\end{tabular}}
\end{center}
\vspace{-0.6cm}
\end{wraptable} 

\vspace{0.1cm}
\noindent \textbf{Ablation on the loss terms}: We have conducted multiple experiments with our proposed model, \textsc{GOPro}, using various loss terms, as presented in Table \ref{tab:loss}. The visual contrastive loss, denoted as $\mathbf{L}_{Con}$, is typically utilized to reduce the difference between two different self-supervised views of augmented image features from MoCo v3. Discarding this loss results in a decrease in performance by almost 1.35\%.

Furthermore, the employment of $\mathbf{L}_{Sem}$ enhances the efficacies of the semantic space, leading to an additional improvement in the results by 2.2\%. We observe that \textsc{GOPro} experiences a decline in performance by 1.92\% and 3\% for single augmentations with MoCo v3 ($x_1$) and AugMix ($x_2$), respectively when compared to the full \textsc{GOPro} model. These findings highlight the importance of both losses, along with $\mathbf{L}_{ce}$, which are responsible for the improved performance of \textsc{GOPro}. 

% \footnote{More ablations, model complexity analysis, and visualizations are available in the Supplementary material.}

\section{Takeaways}
% \vspace{-0.3cm}
In this paper, we present a comprehensive analysis of how self-supervised learning can enhance vision-language models. We propose a novel approach called \textsc{GOPro} that ensures consistency among the augmented views of input images in both the visual and semantic space of CLIP, using innovative loss functions. Furthermore, we introduce a new prompt learning framework in \textsc{GOPro} that leverages visual features by disentangling content and style information and incorporates them into prompt learning through a learnable encoder-decoder-based text projector. Our experimental results demonstrate that \textsc{GOPro} outperforms benchmark prompting methods in three challenging domain generalization tasks involving class, domain, and dataset shifts. Additionally, we are excited to explore the potential of \textsc{GOPro} for more specific applications, such as medical imaging and remote sensing, among others, in the future.

\bibliography{egbib}

\vspace{-0.1cm}
\section{Additional Results}
\label{results}
\noindent\textbf{Base-to-New (B2N) class generalization in details:} In Table \ref{tab:B2N_details}, we have shown the detailed results of our proposed \textsc{GOPro} and other prompting techniques on 11 datasets for B2N generalization task. \textsc{GOPro} is very much successful to beat others on each and every datasets while considering the harmonic mean (HM) of base and new classes. It is important to notice that \textsc{GOPro} has shown significant performance in one of the most fine-grained dataset FGVC-Aircraft and outperforms others by at least of 0.38\% of margin. \\

%%%%%%%%%%%%%%%%%%%% 
\begin{table*}[!ht]
 % \vspace{-0.2cm}
\scriptsize{
    \centering
    \caption{\textcolor{royalblue(traditional)}{Comparison with state-of-the-art methods on base-to-new generalization. \textsc{GOPro} shows better generalization performance over existing methods on 11 different recognition datasets on $16$-shots with context length, $\mathcal{M}$=4. HM represents the harmonic mean.}
    \label{tab:B2N_details}}
    \vspace{0.3cm}
    \scalebox{0.915}{
    \begin{tabular}{lccclccclcccc} 
    % \toprule
     \multicolumn{4}{c}{(a) \textbf{Average over 11 datasets}}&\multicolumn{4}{c}{(b) ImageNet}&\multicolumn{4}{c}{(c) Caltech101}\\\cmidrule(lr){1-4}\cmidrule(lr){5-8}\cmidrule(lr){9-12}
     
\cellcolor[gray]{0.9}Method&\multicolumn{1}{c}{Base}&\multicolumn{1}{c|}{New}&\multicolumn{1}{c}{HM}&\cellcolor[gray]{0.9}Method&\multicolumn{1}{c}{Base}&\multicolumn{1}{c|}{New}&\multicolumn{1}{c}{HM}&\cellcolor[gray]{0.9}Method&\multicolumn{1}{c}{Base}&\multicolumn{1}{c|}{New}&\multicolumn{1}{c}{HM}\\\cmidrule(lr){1-4}\cmidrule(lr){5-8}\cmidrule(lr){9-12}

 \cellcolor[gray]{0.9} CLIP \cite{clip}&\multicolumn{1}{c}{ 69.34}&\multicolumn{1}{c|}{ 74.22}&\multicolumn{1}{c}{71.70 }&
 \cellcolor[gray]{0.9}CLIP \cite{clip}&\multicolumn{1}{c}{ 72.43}&\multicolumn{1}{c|}{68.14 }&\multicolumn{1}{c}{70.22 }&
 \cellcolor[gray]{0.9}CLIP \cite{clip}&\multicolumn{1}{c}{96.84 }&\multicolumn{1}{c|}{ 94.00}&\multicolumn{1}{c}{ 95.40}\\
 
  \cellcolor[gray]{0.9} SLIP \cite{slip}&\multicolumn{1}{c}{ 69.77}&\multicolumn{1}{c|}{74.28}&\multicolumn{1}{c}{71.96}&
 \cellcolor[gray]{0.9}SLIP \cite{slip}&\multicolumn{1}{c}{ 72.95}&\multicolumn{1}{c|}{69.76}&\multicolumn{1}{c}{71.32}&
 
 \cellcolor[gray]{0.9}SLIP \cite{slip}&\multicolumn{1}{c}{96.97}&\multicolumn{1}{c|}{94.05}&\multicolumn{1}{c}{95.49}\\
  
    \cellcolor[gray]{0.9}CoOp \cite{coop}&\multicolumn{1}{c}{ 82.69}&\multicolumn{1}{c|}{ 63.22}&\multicolumn{1}{c}{ 71.66}&
    \cellcolor[gray]{0.9}CoOp \cite{coop}&\multicolumn{1}{c}{76.47 }&\multicolumn{1}{c|}{67.88 }&\multicolumn{1}{c}{71.92 }&
    \cellcolor[gray]{0.9}CoOp \cite{coop}&\multicolumn{1}{c}{ 98.00}&\multicolumn{1}{c|}{89.81 }&\multicolumn{1}{c}{ 93.73}\\
\cellcolor[gray]{0.9}CoCoOp \cite{cocoop}&\multicolumn{1}{c}{80.47 }&\multicolumn{1}{c|}{71.69 }&\multicolumn{1}{c}{ 75.83}&
\cellcolor[gray]{0.9}CoCoOp \cite{cocoop}&\multicolumn{1}{c}{ 75.98}&\multicolumn{1}{c|}{70.43 }&\multicolumn{1}{c}{ 73.10}&
\cellcolor[gray]{0.9}CoCoOp \cite{cocoop}&\multicolumn{1}{c}{ 97.96}&\multicolumn{1}{c|}{93.81 }&\multicolumn{1}{c}{95.84 }\\

\cellcolor[gray]{0.9}MaPLe \cite{maple}&\multicolumn{1}{c}{ 82.28}&\multicolumn{1}{c|}{75.14 }&\multicolumn{1}{c}{ 78.55}&
\cellcolor[gray]{0.9}MaPLe \cite{maple}&\multicolumn{1}{c}{76.66 }&\multicolumn{1}{c|}{ 70.54}&\multicolumn{1}{c}{ 73.47}&
\cellcolor[gray]{0.9}MaPLe \cite{maple}&\multicolumn{1}{c}{ 97.74}&\multicolumn{1}{c|}{94.36 }&\multicolumn{1}{c}{96.02 }\\

\cellcolor[gray]{0.9}\textsc{StyLIP} \cite{stylip}& \multicolumn{1}{c}{83.22}&\multicolumn{1}{c|}{75.94 }&\multicolumn{1}{c}{79.41}&
\cellcolor[gray]{0.9}\textsc{StyLIP} \cite{stylip}& \multicolumn{1}{c}{77.15}&\multicolumn{1}{c|}{ 71.34}&\multicolumn{1}{c}{ 74.13}&
\cellcolor[gray]{0.9}\textsc{StyLIP} \cite{stylip}& \multicolumn{1}{c}{ 98.23}&\multicolumn{1}{c|}{94.91 }&\multicolumn{1}{c}{96.54}\\

\cmidrule(lr){1-4}\cmidrule(lr){5-8}\cmidrule(lr){9-12}
%   \cmidrule(lr){1-4}\cmidrule(lr){5-8}\cmidrule(lr){9-12}

\cellcolor{green!15}\textsc{GOPro}&\multicolumn{1}{c}{\textbf{84.21}}&\multicolumn{1}{c|}
{{\textbf{77.32}}}&\multicolumn{1}{c}{\textbf{80.62}}& 

\cellcolor{green!15}\textsc{GOPro}&\multicolumn{1}{c}{\textbf{78.56}}&\multicolumn{1}{c|}{\textbf{73.22}}&\multicolumn{1}{c}
{\textbf{75.80}}&

\cellcolor{green!15}\textsc{GOPro}&\multicolumn{1}{c}
{\textbf{98.86} }&\multicolumn{1}{c|}{\textbf{95.78}}&\multicolumn{1}{c}
{\textbf{97.30}}\\
\cmidrule(lr){1-4}\cmidrule(lr){5-8}\cmidrule(lr){9-12}
%%%%%%%%%%%%%%%%%%%%%%%%%%%%%%%%%%%%%%%%%%%%%%%%%%%%%%%%%%%%%%%%
&&&&&&&&&&&\\
 \multicolumn{4}{c}{(d) OxfordPets}&\multicolumn{4}{c}{(e) StanfordCars}&\multicolumn{4}{c}{(f) Flowers102}\\\cmidrule(lr){1-4}\cmidrule(lr){5-8}\cmidrule(lr){9-12}
     
 \cellcolor[gray]{0.9}Method&\multicolumn{1}{c}{Base}&\multicolumn{1}{c|}{New}&
 \multicolumn{1}{c}{HM}&\cellcolor[gray]{0.9}Method&\multicolumn{1}{c}{Base}&
 \multicolumn{1}{c|}{New}&\multicolumn{1}{c}{HM}&\cellcolor[gray]{0.9}Method&
 \multicolumn{1}{c}{Base}&\multicolumn{1}{c|}{New}&
 \multicolumn{1}{c}{HM}\\\cmidrule(lr){1-4}\cmidrule(lr){5-8}\cmidrule(lr){9-12}

\cellcolor[gray]{0.9}  CLIP \cite{clip}&\multicolumn{1}{c}{ 91.17}&
  \multicolumn{1}{c|}{97.26 }&\multicolumn{1}{c}{ 94.12}&
 \cellcolor[gray]{0.9} CLIP \cite{clip}&\multicolumn{1}{c}{63.37 }&
  \multicolumn{1}{c|}{74.89}&\multicolumn{1}{c}{68.65 }&
\cellcolor[gray]{0.9}  CLIP \cite{clip}&\multicolumn{1}{c}{72.08 }&\multicolumn{1}{c|}{77.80}&\multicolumn{1}{c}{74.83 }\\

  \cellcolor[gray]{0.9}  SLIP \cite{slip}&\multicolumn{1}{c}{ 91.23}&
  \multicolumn{1}{c|}{97.04 }&\multicolumn{1}{c}{ 94.05}&
 \cellcolor[gray]{0.9} SLIP \cite{slip}&\multicolumn{1}{c}{63.52 }&
  \multicolumn{1}{c|}{74.92}&\multicolumn{1}{c}{68.75 }&
\cellcolor[gray]{0.9}  SLIP \cite{slip}&\multicolumn{1}{c}{72.17}&\multicolumn{1}{c|}{\textbf{77.87}}&\multicolumn{1}{c}{74.91 }\\

   \cellcolor[gray]{0.9} CoOp \cite{coop}&\multicolumn{1}{c}{93.67 }&
    \multicolumn{1}{c|}{ 95.29}&\multicolumn{1}{c}{ 94.47}&
  \cellcolor[gray]{0.9}  CoOp \cite{coop}&\multicolumn{1}{c}{ \textbf{78.12}}&
    \multicolumn{1}{c|}{60.40 }&\multicolumn{1}{c}{68.13}&
  \cellcolor[gray]{0.9}  CoOp \cite{coop}&\multicolumn{1}{c}{97.60}& \multicolumn{1}{c|}{ 59.67}&\multicolumn{1}{c}{74.06 }\\
    
\cellcolor[gray]{0.9}CoCoOp \cite{cocoop}&\multicolumn{1}{c}{ 95.20}&
\multicolumn{1}{c|}{ 97.69}&\multicolumn{1}{c}{ 96.43}&
\cellcolor[gray]{0.9}CoCoOp \cite{cocoop}&\multicolumn{1}{c}{70.49 }&
\multicolumn{1}{c|}{ 73.59}&\multicolumn{1}{c}{72.01 }&
\cellcolor[gray]{0.9}CoCoOp \cite{cocoop}&\multicolumn{1}{c}{ 94.87}&
\multicolumn{1}{c|}{71.15 }&\multicolumn{1}{c}{81.71 }\\

\cellcolor[gray]{0.9}MaPLe \cite{maple}&\multicolumn{1}{c}{95.43 }&
\multicolumn{1}{c|}{97.76 }&\multicolumn{1}{c}{ 96.58}&
\cellcolor[gray]{0.9}MaPLe \cite{maple}&\multicolumn{1}{c}{72.94 }&
\multicolumn{1}{c|}{74.00 }&\multicolumn{1}{c}{73.47 }&
\cellcolor[gray]{0.9}MaPLe \cite{maple}&\multicolumn{1}{c}{95.92 }&
\multicolumn{1}{c|}{72.46 }&\multicolumn{1}{c}{ 82.56}\\

\cellcolor[gray]{0.9}\textsc{StyLIP} \cite{stylip}& \multicolumn{1}{c}{ 95.96}&\multicolumn{1}{c|}{98.14 }&\multicolumn{1}{c}{97.04}&
\cellcolor[gray]{0.9}\textsc{StyLIP} \cite{stylip}& \multicolumn{1}{c}{75.19}&\multicolumn{1}{c|}{ 74.46}&\multicolumn{1}{c}{74.82}&
\cellcolor[gray]{0.9}\textsc{StyLIP} \cite{stylip}& \multicolumn{1}{c}{ 96.54}&\multicolumn{1}{c|}{73.08}&\multicolumn{1}{c}{83.19}\\

\cmidrule(lr){1-4}\cmidrule(lr){5-8}\cmidrule(lr){9-12}
%   \cmidrule(lr){1-4}\cmidrule(lr){5-8}\cmidrule(lr){9-12}

\cellcolor{green!15}\textsc{GOPro}&\multicolumn{1}{c}{ \textbf{96.36}}&\multicolumn{1}{c|}
{\textbf{98.49} }&\multicolumn{1}{c}{ \textbf{97.41}}&
\cellcolor{green!15}\textsc{GOPro}&\multicolumn{1}{c}
{77.59}&\multicolumn{1}{c|}
{\textbf{75.35}}&\multicolumn{1}{c}
{\textbf{76.45} }&
\cellcolor{green!15}\textsc{GOPro}&\multicolumn{1}{c}
{\textbf{97.73}}&\multicolumn{1}{c|}
{77.91}&\multicolumn{1}{c}
{\textbf{86.70} }\\
\cmidrule(lr){1-4}\cmidrule(lr){5-8}\cmidrule(lr){9-12}
%%%%%%%%%%%%%%%%%%%%%%%%%%%%%%%%%%%%%%%%%%%%%%%%%%%%%%%%%%%%%%%%%
&&&&&&&&&&&\\
 \multicolumn{4}{c}{(g) Food101}&\multicolumn{4}{c}{(h) FGVCAircraft}&\multicolumn{4}{c}{(i) SUN397}\\\cmidrule(lr){1-4}\cmidrule(lr){5-8}\cmidrule(lr){9-12}
     
\cellcolor[gray]{0.9}Method&\multicolumn{1}{c}{Base}&\multicolumn{1}{c|}{New}
 &\multicolumn{1}{c}{HM}&\cellcolor[gray]{0.9}Method&\multicolumn{1}{c}{Base}
 &\multicolumn{1}{c|}{New}&\multicolumn{1}{c}{HM}
 &\cellcolor[gray]{0.9}Method&\multicolumn{1}{c}{Base}&\multicolumn{1}{c|}{New}
 &\multicolumn{1}{c}{HM}\\
 \cmidrule(lr){1-4}\cmidrule(lr){5-8}\cmidrule(lr){9-12}

  \cellcolor[gray]{0.9}CLIP \cite{clip}&\multicolumn{1}{c}{90.10 }&
  \multicolumn{1}{c|}{ 91.22}&\multicolumn{1}{c}{90.66 }&
 \cellcolor[gray]{0.9} CLIP \cite{clip}&\multicolumn{1}{c}{ 27.19}&
  \multicolumn{1}{c|}{36.29}&\multicolumn{1}{c}{31.09 }&
\cellcolor[gray]{0.9}  CLIP \cite{clip}&\multicolumn{1}{c}{ 69.36}&
  \multicolumn{1}{c|}{ 75.35}&\multicolumn{1}{c}{ 72.23}\\

    \cellcolor[gray]{0.9}SLIP \cite{slip}&\multicolumn{1}{c}{90.14}&
  \multicolumn{1}{c|}{ 91.27}&\multicolumn{1}{c}{90.70}&
 \cellcolor[gray]{0.9} SLIP \cite{slip}&\multicolumn{1}{c}{27.49}& \multicolumn{1}{c|}{36.11}&\multicolumn{1}{c}{31.22}&
\cellcolor[gray]{0.9}  SLIP \cite{slip}&\multicolumn{1}{c}{ 69.35}& \multicolumn{1}{c|}{75.39}&\multicolumn{1}{c}{72.24}\\
  
  \cellcolor[gray]{0.9}  CoOp \cite{coop}&\multicolumn{1}{c}{ 88.33}&
    \multicolumn{1}{c|}{ 82.26}&\multicolumn{1}{c}{ 85.19}&
  \cellcolor[gray]{0.9}  CoOp \cite{coop}&\multicolumn{1}{c}{ 40.44}&
    \multicolumn{1}{c|}{22.30 }&\multicolumn{1}{c}{28.75 }&
  \cellcolor[gray]{0.9}  CoOp \cite{coop}&\multicolumn{1}{c}{80.60 }&
    \multicolumn{1}{c|}{65.89 }&\multicolumn{1}{c}{ 72.51}\\
    
\cellcolor[gray]{0.9}CoCoOp \cite{cocoop}&\multicolumn{1}{c}{90.70 }&
\multicolumn{1}{c|}{ 91.29}&\multicolumn{1}{c}{90.99}&
\cellcolor[gray]{0.9}CoCoOp \cite{cocoop}&\multicolumn{1}{c}{ 33.41}&
\multicolumn{1}{c|}{ 23.71}&\multicolumn{1}{c}{27.74 }&
\cellcolor[gray]{0.9}CoCoOp \cite{cocoop}&\multicolumn{1}{c}{79.74 }&
\multicolumn{1}{c|}{ 76.86}&\multicolumn{1}{c}{ 78.27}\\

\cellcolor[gray]{0.9}MaPLe \cite{maple}&\multicolumn{1}{c}{90.71 }&
\multicolumn{1}{c|}{ 92.05}&\multicolumn{1}{c}{91.38 }&
\cellcolor[gray]{0.9}MaPLe \cite{maple}&\multicolumn{1}{c}{37.44 }&
\multicolumn{1}{c|}{35.61 }&\multicolumn{1}{c}{ 36.50}&
\cellcolor[gray]{0.9}MaPLe \cite{maple}&\multicolumn{1}{c}{ 80.82}&
\multicolumn{1}{c|}{78.70 }&\multicolumn{1}{c}{ 79.75}\\

\cellcolor[gray]{0.9}\textsc{StyLIP} \cite{stylip}&\multicolumn{1}{c}{91.20}&\multicolumn{1}{c|}{92.48}&\multicolumn{1}{c}{91.84}&
\cellcolor[gray]{0.9}\textsc{StyLIP} \cite{stylip}&\multicolumn{1}{c}{37.65}&\multicolumn{1}{c|}{35.93}&\multicolumn{1}{c}{36.77}&
\cellcolor[gray]{0.9}\textsc{StyLIP} \cite{stylip}&\multicolumn{1}{c}{\textbf{82.12}}&\multicolumn{1}{c|}{79.95}&\multicolumn{1}{c}{81.02}\\

\cmidrule(lr){1-4}\cmidrule(lr){5-8}\cmidrule(lr){9-12}
%   \cmidrule(lr){1-4}\cmidrule(lr){5-8}\cmidrule(lr){9-12}

\cellcolor{green!15}\textsc{GOPro}&
\multicolumn{1}{c}{\textbf{92.37} }&
\multicolumn{1}{c|}{ \textbf{93.56}}&
\multicolumn{1}{c}{\textbf{92.96}}&
\cellcolor{green!15}\textsc{GOPro}&
\multicolumn{1}{c}{\textbf{37.89}}&
\multicolumn{1}{c|}{\textbf{36.44}}&
\multicolumn{1}{c}{\textbf{37.15}}&
\cellcolor{green!15}\textsc{GOPro}&
\multicolumn{1}{c}{81.94}&
\multicolumn{1}{c|}{\textbf{81.64} }&
\multicolumn{1}{c}{\textbf{81.79}}\\
\cmidrule(lr){1-4}\cmidrule(lr){5-8}\cmidrule(lr){9-12}
%%%%%%%%%%%%%%%%%%%%%%%%%%%%%%%%%%%%%%%%%%%%%%%%%%%%%%%%%%%%%%%
&&&&&&&&&&&&\\
 \multicolumn{4}{c}{(j) DTD}&\multicolumn{4}{c}{(k) EuroSAT}&\multicolumn{4}{c}{(l) UCF101}\\\cmidrule(lr){1-4}\cmidrule(lr){5-8}\cmidrule(lr){9-12}
     
\cellcolor[gray]{0.9}Method&\multicolumn{1}{c}{Base}&\multicolumn{1}{c|}{New}&
 \multicolumn{1}{c}{HM}&\cellcolor[gray]{0.9}Method&\multicolumn{1}{c}{Base}&
 \multicolumn{1}{c|}{New}&\multicolumn{1}{c}{HM}&\cellcolor[gray]{0.9}Method&
 \multicolumn{1}{c}{Base}&\multicolumn{1}{c|}{New}&
 \multicolumn{1}{c}{HM}\\
 \cmidrule(lr){1-4}\cmidrule(lr){5-8}\cmidrule(lr){9-12}

\cellcolor[gray]{0.9}  CLIP \cite{clip}&\multicolumn{1}{c}{ 53.24}&
  \multicolumn{1}{c|}{ 59.90}&\multicolumn{1}{c}{56.37 }&
 \cellcolor[gray]{0.9} CLIP \cite{clip}&\multicolumn{1}{c}{ 56.48}&
  \multicolumn{1}{c|}{ 64.05}&\multicolumn{1}{c}{60.03}&
 \cellcolor[gray]{0.9} CLIP \cite{clip}&\multicolumn{1}{c}{70.53 }&
  \multicolumn{1}{c|}{77.50 }&\multicolumn{1}{c}{73.85 }\\

  \cellcolor[gray]{0.9}  SLIP \cite{slip}&\multicolumn{1}{c}{ 56.71}&
  \multicolumn{1}{c|}{59.30}&\multicolumn{1}{c}{57.98}&
 \cellcolor[gray]{0.9} SLIP \cite{slip}&\multicolumn{1}{c}{ 56.43}&
  \multicolumn{1}{c|}{63.79}&\multicolumn{1}{c}{59.88}&
 \cellcolor[gray]{0.9} SLIP \cite{slip}&\multicolumn{1}{c}{70.55}&
  \multicolumn{1}{c|}{77.56}&\multicolumn{1}{c}{73.89 }\\
  
    \cellcolor[gray]{0.9}CoOp \cite{coop}&\multicolumn{1}{c}{79.44 }&
    \multicolumn{1}{c|}{41.18 }&\multicolumn{1}{c}{54.24 }&
   \cellcolor[gray]{0.9} CoOp \cite{coop}&\multicolumn{1}{c}{92.19 }&
    \multicolumn{1}{c|}{ 54.74}&\multicolumn{1}{c}{ 68.69}&
   \cellcolor[gray]{0.9} CoOp \cite{coop}&\multicolumn{1}{c}{ 84.69}&
    \multicolumn{1}{c|}{ 56.05}&
    \multicolumn{1}{c}{67.46 }\\
    
\cellcolor[gray]{0.9}CoCoOp \cite{cocoop}&\multicolumn{1}{c}{ 77.01}&
\multicolumn{1}{c|}{56.00 }&\multicolumn{1}{c}{64.85 }&
\cellcolor[gray]{0.9}CoCoOp \cite{cocoop}&\multicolumn{1}{c}{87.49 }&
\multicolumn{1}{c|}{ 60.04}&\multicolumn{1}{c}{71.21 }&
\cellcolor[gray]{0.9}CoCoOp \cite{cocoop}&\multicolumn{1}{c}{ 82.33}&
\multicolumn{1}{c|}{ 73.45}&\multicolumn{1}{c}{77.64 }\\

% \cellcolor[gray]{0.9}SVL-Adapter \cite{svl-adapter}&\multicolumn{1}{c}{70.30}&\multicolumn{1}{c|}{59.47 }&\multicolumn{1}{c}{64.43}&

% \cellcolor[gray]{0.9}SVL-Adapter \cite{svl-adapter}&\multicolumn{1}{c}{84.61}&\multicolumn{1}{c|}{67.46 }&\multicolumn{1}{c}{75.10}&

% \cellcolor[gray]{0.9}SVL-Adapter \cite{svl-adapter}&\multicolumn{1}{c}{81.15}&\multicolumn{1}{c|}{74.37 }&\multicolumn{1}{c}{77.61}\\

\cellcolor[gray]{0.9}MaPLe \cite{maple}&\multicolumn{1}{c}{ 80.36}&
\multicolumn{1}{c|}{ 59.18}&\multicolumn{1}{c}{68.16 }&
\cellcolor[gray]{0.9}MaPLe \cite{maple}&\multicolumn{1}{c}{94.07 }&
\multicolumn{1}{c|}{73.23 }&\multicolumn{1}{c}{ 82.35}&
\cellcolor[gray]{0.9}MaPLe \cite{maple}&\multicolumn{1}{c}{ 83.00}&
\multicolumn{1}{c|}{78.66 }&\multicolumn{1}{c}{80.77 }\\

\cellcolor[gray]{0.9}\textsc{StyLIP} \cite{stylip}&\multicolumn{1}{c}{81.57}&\multicolumn{1}{c|}{61.72}&\multicolumn{1}{c}{70.27}&
\cellcolor[gray]{0.9}\textsc{StyLIP} \cite{stylip}&\multicolumn{1}{c}{94.61}&\multicolumn{1}{c|}{ 74.06}&\multicolumn{1}{c}{83.08}&
\cellcolor[gray]{0.9}\textsc{StyLIP} \cite{stylip}&\multicolumn{1}{c}{85.19}&\multicolumn{1}{c|}{\textbf{79.22}}&\multicolumn{1}{c}{82.10}\\

\cmidrule(lr){1-4}\cmidrule(lr){5-8}\cmidrule(lr){9-12}
%   \cmidrule(lr){1-4}\cmidrule(lr){5-8}\cmidrule(lr){9-12}

\cellcolor{green!15}\textsc{GOPro}&\multicolumn{1}{c}{ \textbf{82.41}}&\multicolumn{1}{c|}{\textbf{62.95}}&\multicolumn{1}{c}{ \textbf{71.38}}&\cellcolor{green!15}\textsc{GOPro}&\multicolumn{1}{c}{\textbf{94.92}}&\multicolumn{1}{c|}{\textbf{76.27}}&\multicolumn{1}{c}{\textbf{84.58}}&\cellcolor{green!15}\textsc{GOPro}&\multicolumn{1}{c}{\textbf{87.67}}&\multicolumn{1}{c|}{78.91}&\multicolumn{1}{c}{\textbf{83.06}}\\

\cmidrule(lr){1-4}\cmidrule(lr){5-8}\cmidrule(lr){9-12}
    \end{tabular}}}
\end{table*}

\noindent\textbf{Sensitivity to the variation in the number of shots:}
We evaluate the performance of our proposed \textsc{GOPro} on the base-to-new class generalization task, varying the number of shots from 1 to 16 and taking all training samples from every base class. Table \ref{tab:shots} compares our results with state-of-the-art prompting techniques. For this evaluation, we use a context length ($\mathcal{M}$) of 4 and ViT-B/16 as the visual feature backbone, by placing the class token at the end and utilizing a unified context vector. Since CLIP is a zero-shot approach, we exclude it and focus on few-shot-based prompting methods. We present our results as harmonic mean (HM) of base and new classes on average over 11 datasets. Our \textsc{GOPro} consistently outperforms benchmark prompt learning-based methods by a minimum margin of 0.2\%, 0.5\%, 2.3\%, 1.2\% and 1.2\% for 1, 2, 4, 8, 16 shots and all training samples, respectively. \\

\begin{figure}
    \centering
    \includegraphics[scale=0.45]{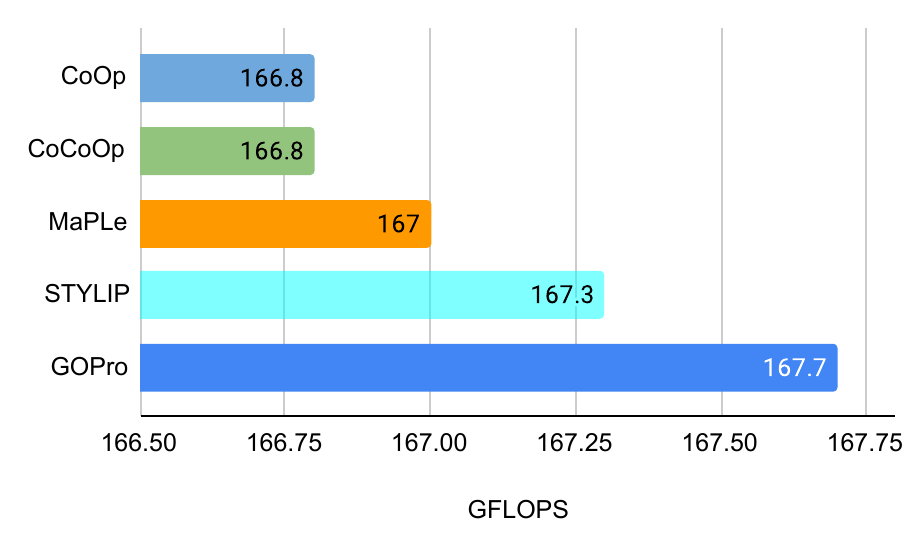}
    % \vspace{0.045cm}
    \caption{\textcolor{royalblue(traditional)}{Comparison of the computational complexity of \textsc{GOPro} among different prompting methods in terms of GFLOPS.}}
    \label{fig:gflops}
    %\vspace{-0.6cm}
\end{figure}

\begin{table*}[!ht]
% \vspace{-0.3cm}
% \scriptsize{
    \centering
    \caption{\textcolor{royalblue(traditional)}{Comparison of \textsc{GOPro} with state-of-the-art methods on varying the number of shots for the B2N class generalization task on average of 11 datasets. We choose harmonic mean (H) of base and new classes for comparison, as well as to depict the generalization trade-off.}\label{tab:shots}}
\scalebox{0.8}{
\begin{tabular}{lccccccc}\\
\toprule
\rowcolor{gray!20} {Method} & {1-shot} & {2-shot} & {4-shot} & {8-shot} & {16-shot} & {All} \\ 
        \midrule
      \cellcolor[gray]{0.9}CoOp &67.14 &67.32 &68.28 &69.33 &71.66 &\boxit{0.36in}71.89  \\

      \cellcolor[gray]{0.9}CoCoOp &70.67 &71.94 &72.45 &74.22 &\boxit{0.36in}75.83 &75.36 \\

      \cellcolor[gray]{0.9}\textsc{StyLIP} &73.98 &74.46 &75.57 &78.86 &\boxit{0.36in}79.41 &79.25 \\

      \cellcolor{green!15}\textsc{GOPro} &74.17 &74.95 &77.84 &79.31 &\textbf{80.62} &80.48 \\

\bottomrule
\end{tabular}}
\end{table*}

\noindent\textbf{Sensitivity to the prompt initialization strategy:}
In Table \ref{tab:initialization}, we examine the effectiveness of three distinct prompt initialization strategies for single-source multi-target (SSMT) domain generalization. The results emphasize that manual initialization using \texttt{"a photo of a"} surpasses random initialization and no initialization strategies significantly for all the target datasets, except ImageNetV2 and ImageNet-R. However, manual initialization outperforms other strategies in the evaluation of the source domain i.e. ImageNet dataset.\\

\begin{table*}[!ht]
% \vspace{-0.3cm}
% \scriptsize{
    \centering
    \caption{\textcolor{royalblue(traditional)}{Comparison of \textsc{GOPro} with the benchmark prompt initialization methods for domain generalization across datasets. We train the model on ImageNet using $16$-shots with CLIP ViT-B/16 and test on $4$ other datasets.}}
    \vspace{0.4cm}
    \scalebox{0.65}{
    \begin{tabular}{lcccccc}
    \toprule
       \rowcolor{gray!20} & \multicolumn{1}{c}{\textbf{Source}} & \multicolumn{4}{c}{\textbf{Target}}  \\ \cmidrule(lr){2-2} \cmidrule(lr){3-6}
        
       \rowcolor{gray!20} \multirow{-2}{*}{\textbf{Method}} & {ImageNet} & {ImageNetV2} & {ImageNet-Sketch} & {ImageNet-A} & {ImageNet-R} \\ 
        \midrule
         \cellcolor[gray]{0.9} random initilization & 72.89 & \textbf{66.03} & 49.34 & 51.96 & \textbf{78.24}  \\ 
         \cellcolor[gray]{0.9} without initialization & 71.56& 63.44& 49.10 & 48.62& 77.92 \\
         \cellcolor[gray]{0.9} manual initialization &\textbf{73.27} &65.35 & \textbf{50.36}&	\textbf{52.23}&	78.02\\
\bottomrule
        % GOPro&  &  & & & & & & & & & & \\\hline
        
    \end{tabular}}
    \label{tab:initialization}
\end{table*}

\noindent\textbf{Computational Complexity:} We run our model on NVIDIA RTX A6000 GPU with 48 GB card. Fig. \ref{fig:gflops} represents the comparison of computational complexity between different prompting techniques (CoOp \cite{coop}, CoCoOp \cite{cocoop}, MaPLe \cite{maple} and \textsc{StyLIP} \cite{stylip}) in terms of GFLOPS. MaPLe and \textsc{StyLIP} require 0.12\% and 0.3\% more computational overhead than CoCoOp respectively, whereas \textsc{GOPro} needs 0.24\%, 0.42\% and 0.54\% more resources than \textsc{StyLIP}, MaPLe and CoCoOp. However, \textsc{GOPro} outperforms well the state-of-the-art methods on all of the three generalization tasks i.e. base-to-new, cross-dataset and domain generalization by smart margins.
\end{document}